\documentclass[final,5p,times,authoryear,twocolumn]{elsarticle}
\usepackage[colorlinks, %%改变引用颜色
            linkcolor=blue,  
            anchorcolor=blue, 
            citecolor=blue,
            ]{hyperref}
\usepackage{graphicx}
\usepackage{subfigure}
\journal{NEURAL NETWORKS}
\usepackage{booktabs}
\usepackage{caption}
\usepackage{comment}
\usepackage{caption}
\usepackage{titlesec}
\titleformat{\subsection}
{\normalfont\bfseries}{\textbf{\thesubsection}}{1em}{\textbf}
\titleformat{\subsubsection}
{\normalfont\bfseries}{\textbf{\thesubsubsection}}{1em}{\textbf}
\usepackage{float}
\usepackage{afterpage}
\usepackage{amsmath}
\usepackage[T1]{fontenc}

\begin{document}
\begin{frontmatter}

\title{Dual Pipeline Style Transfer with Input Distribution Differentiation}
\author{ShiQi Jiang,JunJie Kang,YuJian Li\affiliation{organization={Guilin University of Electronic Technology},
            city={Guilin},
            country={China}}}

\begin{abstract}
The color and texture dual pipeline architecture (\textbf{CTDP}) suppresses texture representation and artifacts through masked total variation loss (\textbf{Mtv}), and further experiments have shown that smooth input can almost completely eliminate texture representation. We have demonstrated through experiments that smooth input is not the key reason for removing texture representations, but rather the distribution differentiation of the training dataset. Based on this, we propose an input distribution differentiation training strategy (\textbf{IDD}), which forces the generation of textures to be completely dependent on the noise distribution, while the smooth distribution will not produce textures at all. Overall, our proposed distribution differentiation training strategy allows for two pre-defined input distributions to be responsible for two generation tasks, with noise distribution responsible for texture generation and smooth distribution responsible for color smooth transfer. Finally, we choose a smooth distribution as the input for the forward inference stage to completely eliminate texture representations and artifacts in color transfer tasks.
\end{abstract}
\afterpage{
	\begin{figure*}[t]
		\centering
		\includegraphics[width=\textwidth,height=\textheight,keepaspectratio]{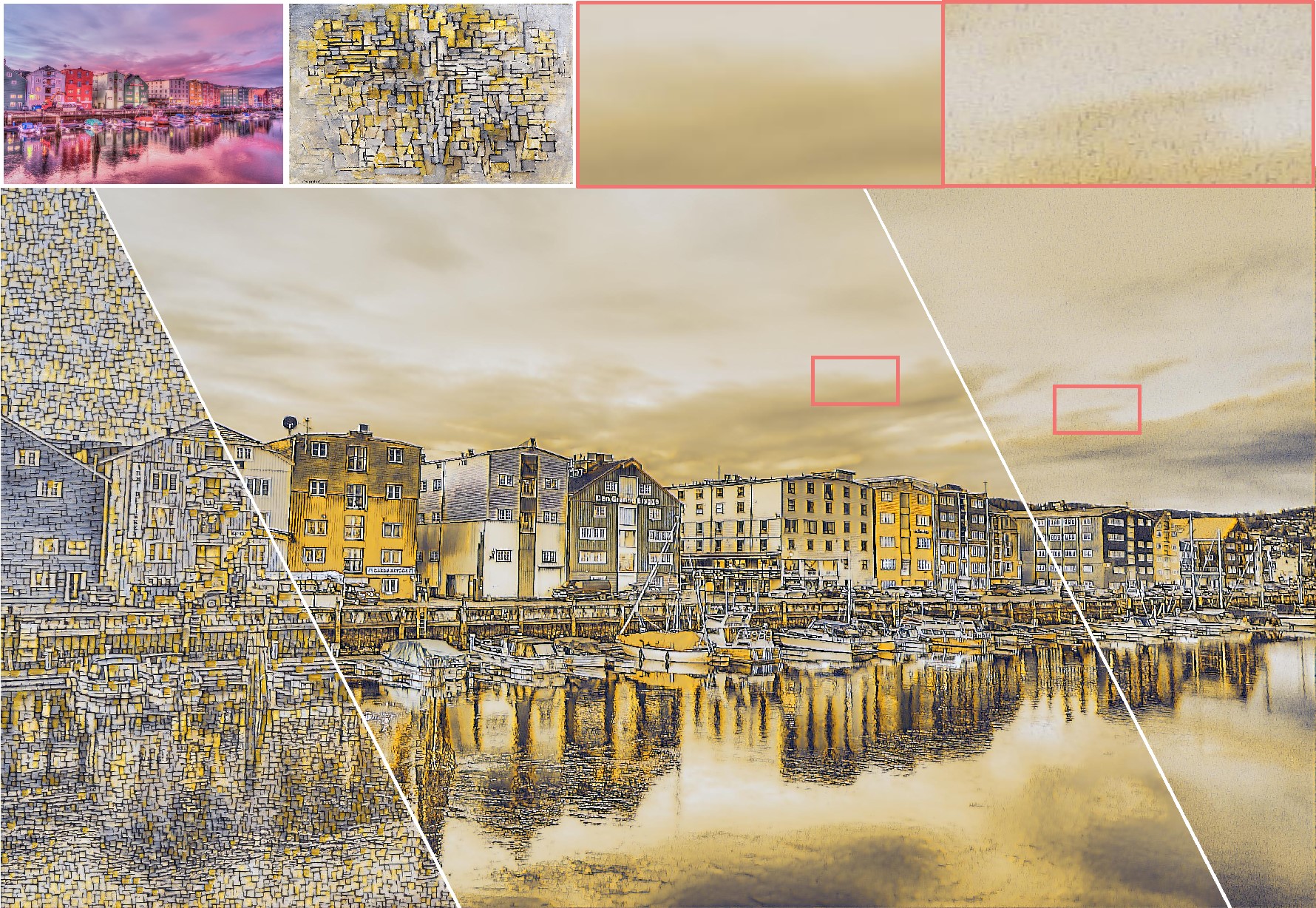}
		\caption{\textbf{The 4K super-resolution stylized image} generated by our proposed IDD. The top of the image displays a content image, a style image, and an enlarged result in the red box area of the stylized result. At the bottom of the image are three stylized results that are concatenated, namely texture transfer results, color transfer prediction results of smooth input, and color transfer prediction results of noise added after smoothing.}
		\label{4ktupian}
	\end{figure*}
}

\end{frontmatter}

\section{Introduction}
Style transfer is a highly attractive image processing technique that can transfer the unique colors and texture styles of artworks to content images. In recent years, methods for style transfer have been widely proposed, which can be roughly divided into two categories: online image optimization and model optimization.

The representative of image optimization methods is (\cite{gatys}), which innovatively transfers gradients to the input image and iteratively optimizes the input content image directly. The style pattern is represented by the feature correlation of deep convolutional neural networks (VGG, \cite{vgg}). Subsequent work mainly focuses on different forms of loss functions (\cite{15,27}). However, this slow online optimization method has a high time cost and greatly reduces its actual citation value. In contrast, the model optimization method effectively solves the time-consuming problem of online iteration through offline model training and forward reasoning. There are three main types of model optimization: (1) Training exclusive style transformation models for a single artistic style (\cite{Perceptual,18,32,33}) Synthesize stylized images using a single given artistic style image; (2) Training model that can convert multiple styles (\cite{3,7,35,20,37}) Introducing various network architectures while handling multiple styles; (3) Arbitrary style transformation model (\cite{12,21,microast,collaborative,meta,DynamicIN}) used different mechanisms such as feature modulation and matching to transfer any artistic style.

Reviewing all the methods mentioned above, only DcDae (\cite{DcDae}) and CTDP (\cite{CTDP}) can simultaneously accomplish the tasks of color and texture transfer. The color transfer results obtained by DcDae are byproducts of direct decoding in shallow layers, while CTDP yields high-quality color transfer results with constraints, effectively suppressing the texture representation from the reference style in the color transfer task. CTDP has been demonstrated to efficiently and rapidly achieve high-quality color and texture transfer simultaneously. Our primary focus is to completely eliminate texture representation in the color transfer branch rather than merely suppressing it.

\begin{figure*}[t]
	\centering
	\includegraphics[width=\textwidth,height=\textheight,keepaspectratio]{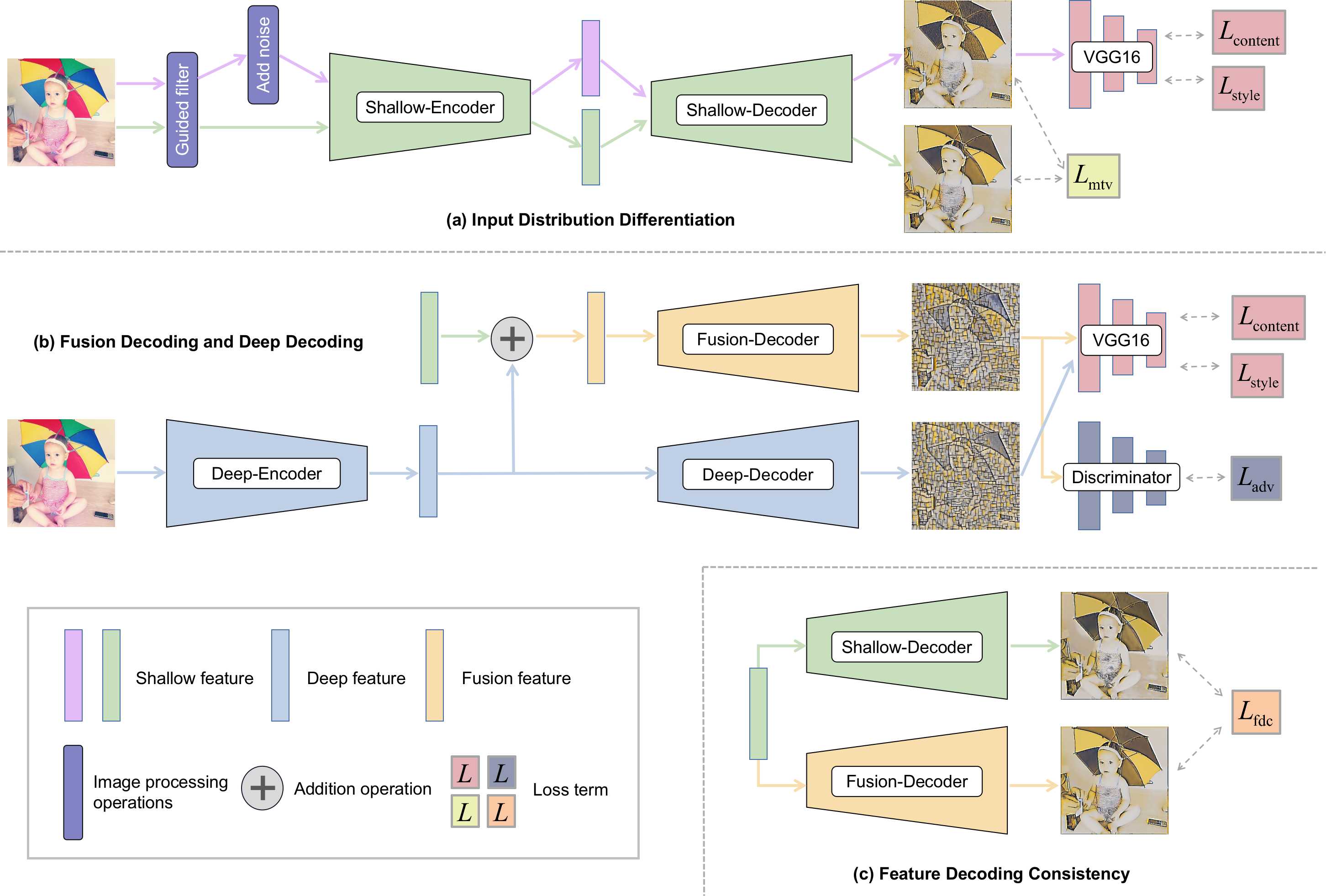}
	\caption{Architecture illustration of the proposed \textbf{IDD}. See Section 3 for details.}
	\label{jiegou}
\end{figure*}

CTDP asserts that decoupling and separating color information within the Gram (\cite{gatys}) matrix is extremely challenging. Instead, it achieves the color transfer task by suppressing the model's texture generation capabilities and the texture representation of the output. Specifically, CTDP reduces the receptive field of the model and Gram matrix calculations through Branch Style loss and a shallower model structure. Additionally, it further suppresses texture representation in the color transfer results using Masked Total Variation loss. While CTDP is capable of achieving good texture suppression in color transfer images, it's important to note that this approach focuses on suppression rather than complete elimination. The model still retains the ability to generate textures, leading to subtle texture representations in the transfer results, and there is differentiation in texture suppression. In more visual analysis, CTDP posits that texture generation depends entirely on the discontinuity of input images and guided filtering (\cite{guide}) can smooth the input to eliminate texture representation.

In the face of the above problems, we propose a input distribution differentiation training strategy (\textbf{IDD}), which compels the generation of textures to rely entirely on noise distribution, while the smooth distribution will not produce any textures. If the input data is smoothed without adding noise, the model will completely lose its ability to generate texture for such input distribution, and thus achieve the effect of completely eliminating texture representation. Furthermore, if all inputs adhere to the same distribution, it solves the problem of differentiated texture suppression. In comparison to state-of-the-art models, we can simultaneously achieve superior color and texture transfer effects. In summary, our contributions are as follows:
\begin{itemize}
    \item Experimental evidence confirms that guided filtering is not the primary reason for removing texture representations but rather the distribution differences in the training dataset.
    \item We introduce a input distribution differentiation training strategy, compelling texture generation to rely entirely on noise distribution, while smooth distribution will not generate textures at all.
    \item During the inference stage, all inputs are constrained to follow the same smooth distribution, thus addressing the issue of differential performance in texture suppression.
    \item Detailed feature visualization analysis of texture generation mechanism and found that input smoothing operation can almost completely eliminate texture structure representation.
    \item Extensive qualitative and quantitative experiments demonstrate that our approach can rapidly achieve high-quality color and texture style transfer simultaneously, while completely eliminating texture representations in color transfer.
\end{itemize} 

\section{Related work}
\subsection{Neural Style Transfer}
With the groundbreaking work of (\cite{gatys}), the era of neural style transfer (NST) has arrived. The visual appeal of style transfer has inspired subsequent researchers to improve in many aspects, including efficiency (\cite{Perceptual,32}); Quality (\cite{stroke,17,10,xie,DcDae}); Diversity (\cite{wang,chen}) and User Control (\cite{zhang,cham}); Despite significant progress, existing methods still cannot decouple the information represented in the Gram (\cite{gatys}) matrix.

\subsection{Color Style Transfer}
Unlike artistic style transfer (\cite{stroke,17,10,xie,meta,microast,demystifying,SANet}), it usually changes both color and texture structure simultaneously. The purpose of color style transfer (also known as realistic style transfer) is to only transfer colors from one image to another. Traditional methods (\cite{pitie2005n,pitie2007automated,reinhard2001color}) mostly match statistical data of low-level features, such as the mean and variance of images (\cite{reinhard2001color}) or histograms of filter responses (\cite{pitie2005n}). However, if there is a significant appearance difference between the style and the input image, these methods typically transfer unwanted colors. In recent years, many methods for color transfer using convolutional deep learning methods (\cite{photowct2,closed,deep,photorealistic,cap}) have been proposed. For example, (\cite{photorealistic}) Introduced a model with wavelet pooling to reduce distortion. CAP-VSTNet (\cite{cap}) uses a reversible residual network and an unbiased linear transformation module to prevent artifacts. Previous methods have improved in suppressing artifacts and content preservation, but have overlooked the impact of complex textures in reference styles on color transfer. The proposed method solves this problem by reducing receptive fields and masked total variation loss to suppress texture representation in Gram (\cite{gatys}).

\section{Method}

\begin{figure*}[t]
	\centering
	\includegraphics[width=\textwidth,height=\textheight,keepaspectratio]{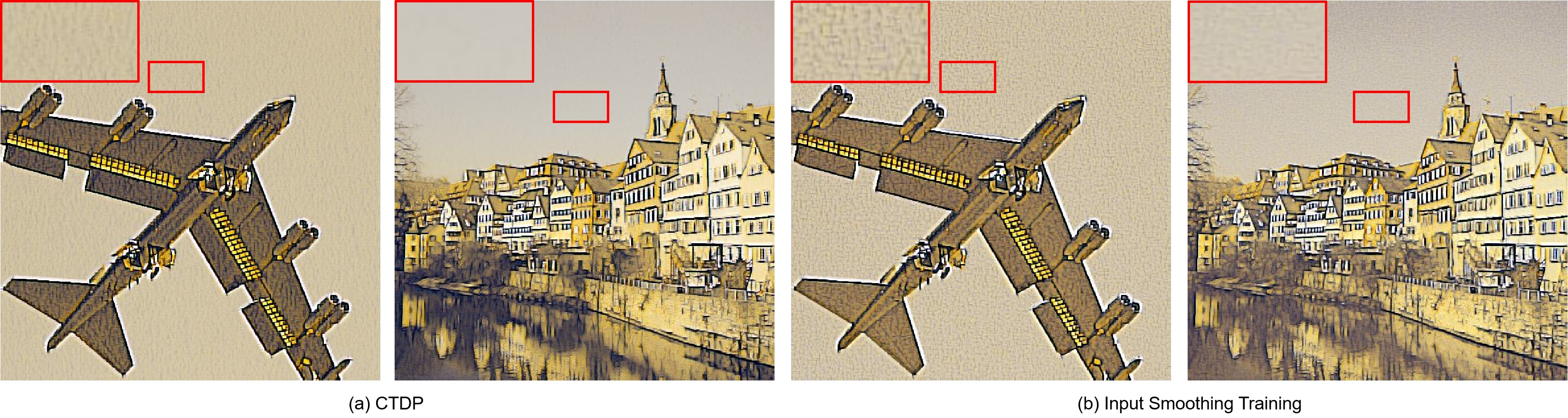}
	\caption{Comparison of prediction results between input noise training scheme and CTDP scheme.}
	\label{Guide}
\end{figure*}

\subsection{Background}
(\cite{CTDP}) pioneered the design of a dual pipeline style transfer (CTDP) framework to simultaneously generate color and texture transfer results, suppressing texture representation in Gram (\cite{gatys}) through masked total variation loss (Mtv).

\textbf{Texture suppression differentiated performance}. CTDP believes that texture differentiation is caused by the continuity of the input image.

\textbf{Input smoothing}. Based on the above assumption, CTDP eliminates all image discontinuities through input smoothing operations to solve the problem of texture suppression differentiation.

\textbf{Feature visualization analysis}. Using only Mtv, noise features are still generated in the feature map, and such noise features will evolve into texture structures after multi-layer convolution. By smoothing the input image, the feature map produces no noise features at all, and ultimately does not produce texture structures.

\textbf{Conclusion}. CTDP believes that input discontinuity will generate noise features in the feature map, and the noise features will evolve into texture structures through convolution operations. Therefore, adopting input smoothing operation can eliminate input discontinuity and feature map noise, and then completely eliminate texture structures.

\subsection{Input Smoothing Training}
Based on the assumption of CTDP, smoothing operations similar to guided filtering (\cite{guide}) an remove texture structures. We attempt to smooth out all training data during the training phase to achieve the effect of removing all texture structures.

As shown in Fig.\ref{Guide}, (a) is the texture suppression differential performance of CTDP, and (b) is the prediction result of our input smoothing training. By comparing the images on the right of (a) and (b), it is evident that, following input smoothing training, even the originally textureless images exhibit the emergence of textures. This experiment demonstrates that smooth inputs will also generate noisy features in the feature maps, ultimately evolving into textured structures, which differs from the assumptions of CTDP. And we can find that (a) the image on the left already produces subtle texture representations in CTDP, which have been accentuated after input smoothing training. Based on the above observations, we formulate the following hypotheses:

(1) Deep learning frameworks are driven by loss functions, and it is impossible to alter the goal of generating textured structures by modifying inputs or intermediate processes;

(2) The distribution differences in the dataset result in differentiated expression of texture suppression. Most images in the dataset follow a discontinuous distribution, so in the model optimization process, texture generation is chosen to model on a discontinuous distribution. Therefore, the continuous distribution of images, like outliers, cannot achieve good stylization effects;

(3) The input smoothing operation is not the fundamental reason for eliminating textures. The texture generation modeling of CTDP is based on the discontinuous distribution of the image, and smoothing operations can precisely eliminate this discontinuous distribution, which indirectly leads to the elimination of texture representation;

\begin{figure}[t]
	\centering
	\includegraphics[width=0.5\textwidth,height=\textheight,keepaspectratio]{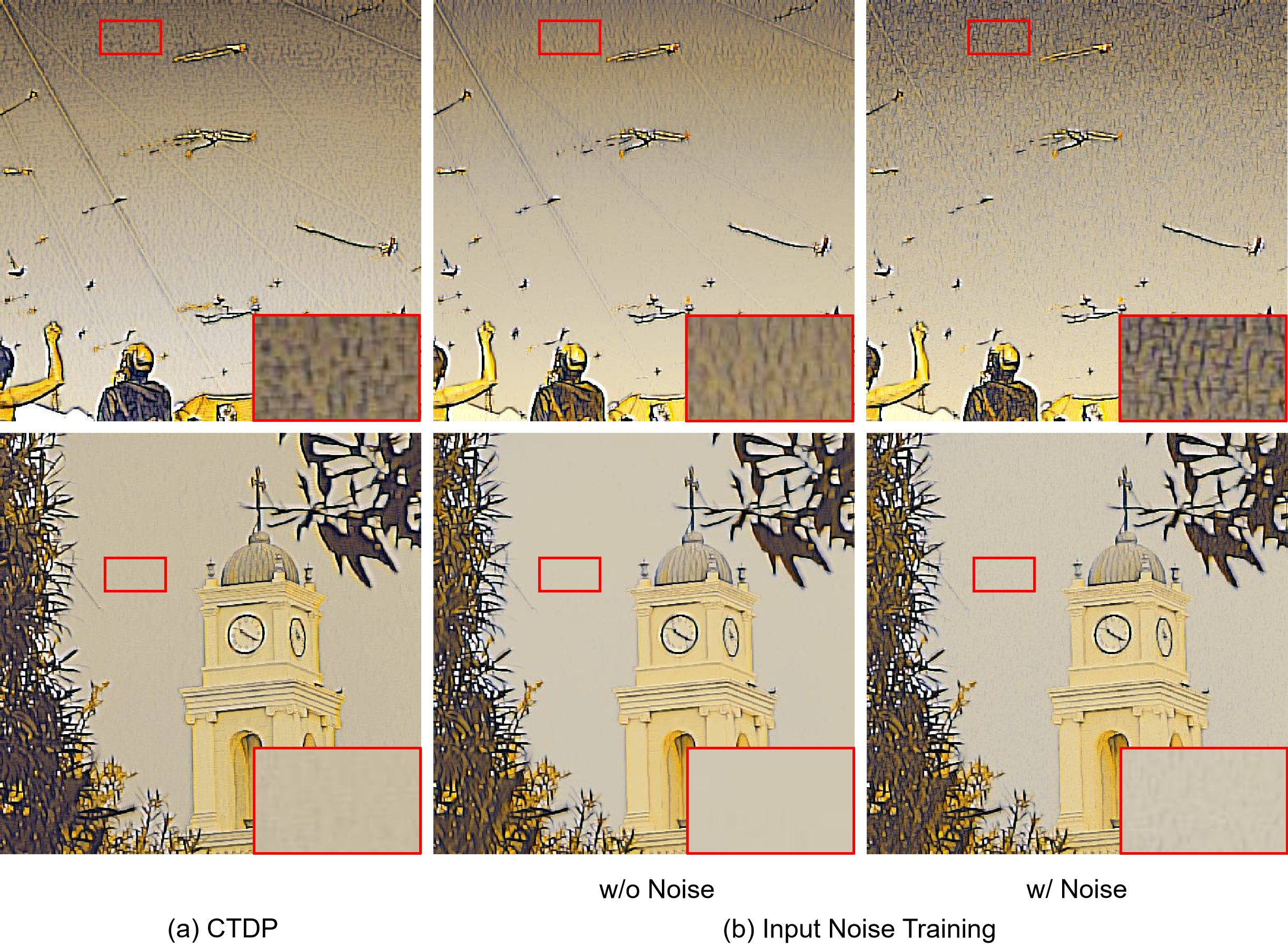}
	\caption{Comparison of prediction results between input smoothing training scheme and CTDP scheme.}
	\label{Noise}
\end{figure}

\subsection{Input Noise Training}
Based on the above experimental assumptions, texture generation in CTDP and input smoothing training is modeled on the discontinuous distribution of the dataset and the smoothed distribution after smoothing operations, respectively. So we can also force texture generation modeling to be based on a predetermined prior distribution. If prior distributions are not added during the inference stage, the model loses its ability to generate textures, thereby achieving the effect of eliminating texture representation.

We attempt to force the model to model texture generation within a predetermined noise distribution (a normal distribution with a mean of 0 and a standard deviation of 0.1). If no noise distribution is added during the inference stage, the model loses its ability to generate texture.

As shown in Fig.\ref{Noise}, (b) is the prediction result of input noise training. On the left is the prediction result without adding noise, and on the right is the prediction result with adding noise. It can be observed that more advanced and complex texture representations were indeed modeled on the noise distribution, and compared to (a), the results without adding noise prediction did indeed have better texture suppression effects. The texture representation was even completely eliminated in the second row of images.

Although (b) exhibits better texture suppression, (b) the image on the left side of the first line indicates that in images with already complex texture representations, it cannot completely eliminate texture representations like in the second line image, which can only further suppress them. We believe that this is due to the inherent discontinuity distribution in the dataset. Under the condition of input noise training, the texture generation modeling of the model is based on a mixed distribution of discontinuity distribution in the original dataset and predetermined noise distribution. And because the modeling of texture generation ability has a stronger dependence on the predetermined noise distribution, without adding noise prediction, better texture suppression effects can be achieved.

\begin{figure}[t]
	\centering
	\includegraphics[width=0.5\textwidth,height=\textheight,keepaspectratio]{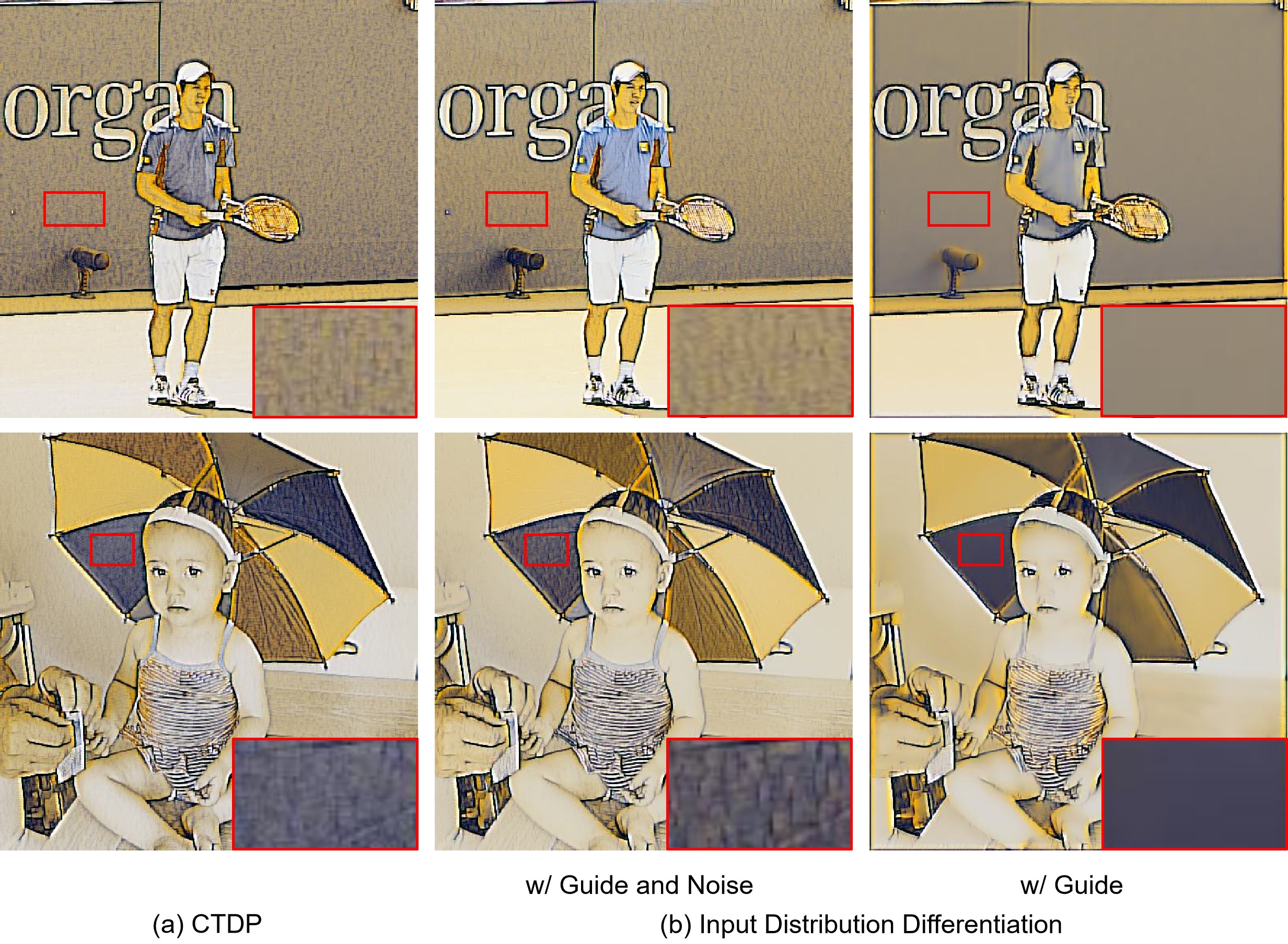}
	\caption{Comparison of prediction results between input distribution differentiation (\textbf{IDD}) scheme and CTDP scheme}
	\label{GuideNoise}
\end{figure}

\subsection{Input Distribution Differentiation}
In order to fully model the texture generation ability of the model on our predetermined noise distribution instead of a mixed distribution that we cannot fully control, we need to ensure that the model input only contains the noise distribution and is not affected by other distributions.

We propose an input distribution differentiation training strategy, which alternates input distribution differentiation training for color transfer branches. Step 1, texture modeling. We smooth the input (guided filtering (\cite{guide})) and then add a predetermined noise distribution to eliminate the inherent discontinuous distribution of the dataset as much as possible to ensure that the input data follows the same noise distribution, forcing the texture generation to be fully modeled on our predetermined noise distribution. Step 2, texture removal. For inputs that are only smoothed, we do not impose branch style loss constraints, and only use the masked total variation loss (Mtv) to further suppress texture representation.

As shown in Fig.\ref{GuideNoise}(b), the left and right images are the results of whether to add noise prediction after smoothing processing. Comparing the left and right images of (b), it was found that the texture generation ability was fully modeled on our predetermined noise distribution. Only by adding noise distribution prediction can the stylized results show texture representation. As long as we do not add noise distribution, our model completely loses the ability to generate texture and produces extremely smooth results.

\section{Experiments}
\subsection{Implementation Details}
\subsection{Comparisons with Prior Arts}

\begin{figure*}[t]
	\centering
	\includegraphics[width=\textwidth,height=\textheight,keepaspectratio]{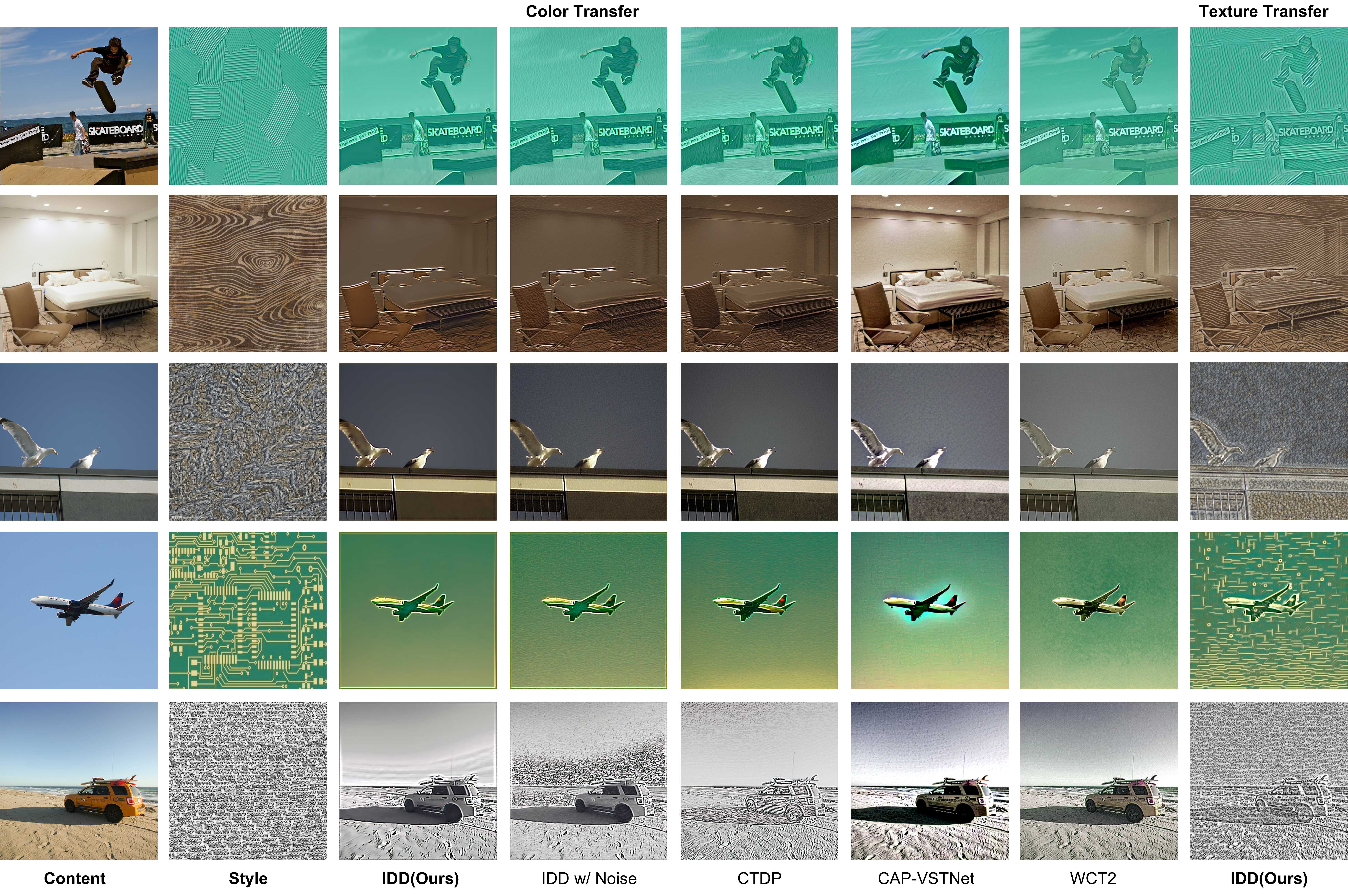}
	\caption{\textbf{Quantitative Comparison} with the state-of-the-art color and texture transfer methods using 1024 resolution input images. Due to the selection of many challenging style images with complex texture structures, it is best to zoom in to better observe artifact suppression and texture structure transfer.}
	\label{duibi}
\end{figure*}
\begin{figure}[t]
	\centering
	\includegraphics[width=0.5\textwidth,height=\textheight,keepaspectratio]{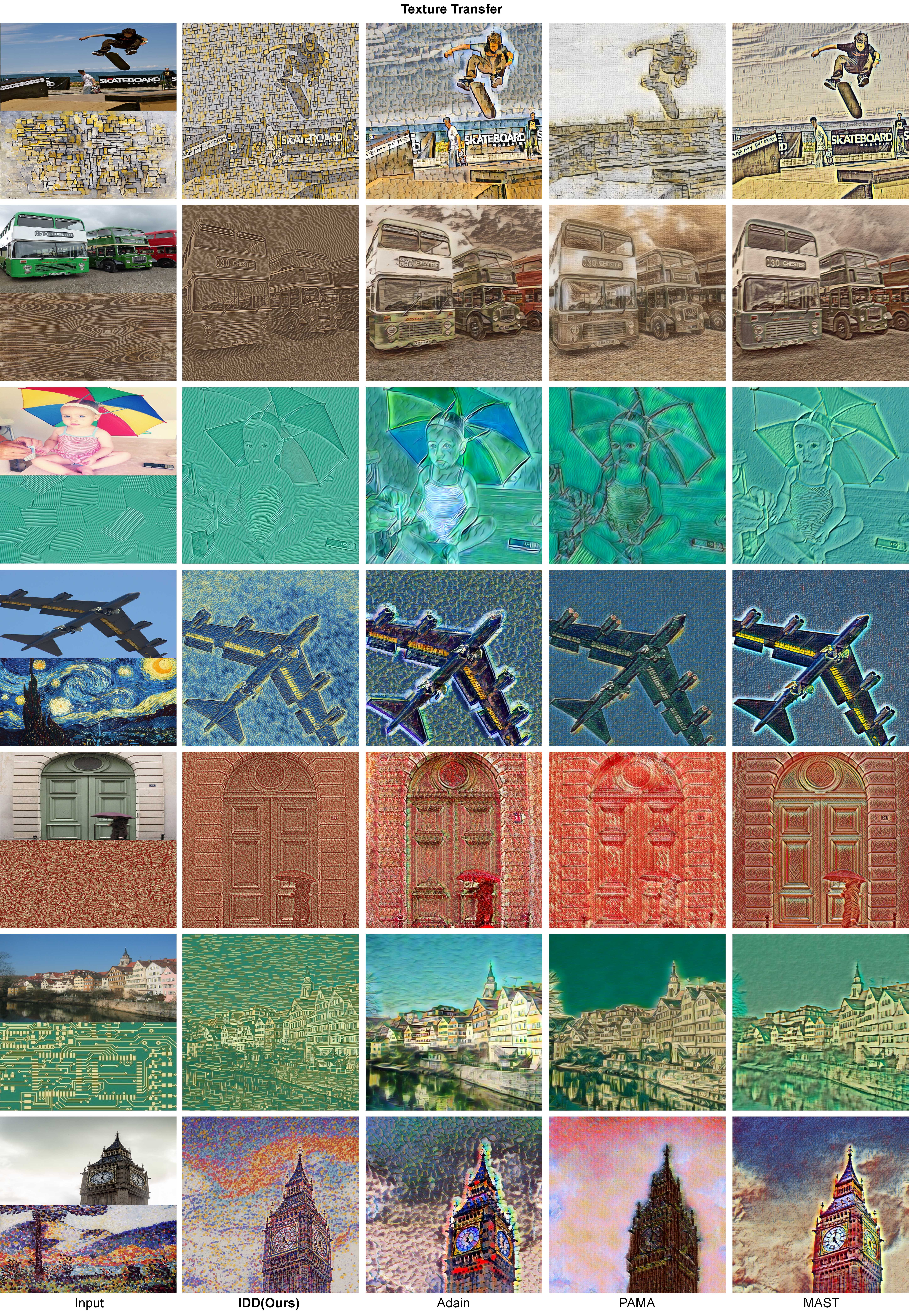}
	\caption{\textbf{Quantitative Comparison} with the state-of-the-art color and texture transfer methods using 1024 resolution input images. Due to the selection of many challenging style images with complex texture structures, it is best to zoom in to better observe artifact suppression and texture structure transfer.}
	\label{duibi}
\end{figure}

\subsubsection{Qualitative Comparison}
\subsection{Ablation Study}
\section{Conclusion}
In this article, we propose an input distribution differentiation training strategy called \textbf{IDD}. This method forces the modeling of texture generation to rely entirely on a predetermined noise distribution, while smooth distribution will not generate texture representation at all. The inference stage ensures that all inputs follow the same smooth distribution, which can completely eliminate texture representations in color transfer branches and solve the problem of differentiated texture suppression. A large number of experiments have proven the effectiveness of this method. Compared to the current level of technology, our IDD is the first model that can completely eliminate the strong texture representation problem in the Gram matrix caused by complex pattern reference images.

\bibliographystyle{elsarticle-harv}
\bibliography{ref}

\begin{thebibliography}{44}
\expandafter\ifx\csname natexlab\endcsname\relax\def\natexlab#1{#1}\fi
\providecommand{\url}[1]{\texttt{#1}}
\providecommand{\href}[2]{#2}
\providecommand{\path}[1]{#1}
\providecommand{\DOIprefix}{doi:}
\providecommand{\ArXivprefix}{arXiv:}
\providecommand{\URLprefix}{URL: }
\providecommand{\Pubmedprefix}{pmid:}
\providecommand{\doi}[1]{\href{http://dx.doi.org/#1}{\path{#1}}}
\providecommand{\Pubmed}[1]{\href{pmid:#1}{\path{#1}}}
\providecommand{\bibinfo}[2]{#2}
\ifx\xfnm\relax \def\xfnm[#1]{\unskip,\space#1}\fi
%Type = Article
\bibitem[{Champandard(2016)}]{cham}
\bibinfo{author}{Champandard, A.J.}, \bibinfo{year}{2016}.
\newblock \bibinfo{title}{Semantic style transfer and turning two-bit doodles
  into fine artworks}.
\newblock \bibinfo{journal}{arXiv preprint arXiv:1603.01768} .
%Type = Inproceedings
\bibitem[{Chen et~al.(2017)Chen, Yuan, Liao, Yu and Hua}]{3}
\bibinfo{author}{Chen, D.}, \bibinfo{author}{Yuan, L.}, \bibinfo{author}{Liao,
  J.}, \bibinfo{author}{Yu, N.}, \bibinfo{author}{Hua, G.},
  \bibinfo{year}{2017}.
\newblock \bibinfo{title}{Stylebank: An explicit representation for neural
  image style transfer}, in: \bibinfo{booktitle}{Proceedings of the IEEE
  conference on computer vision and pattern recognition}, pp.
  \bibinfo{pages}{1897--1906}.
%Type = Inproceedings
\bibitem[{Chen et~al.(2021)Chen, Zhao, Zhang, Wang, Zuo, Li, Xing and
  Lu}]{chen}
\bibinfo{author}{Chen, H.}, \bibinfo{author}{Zhao, L.}, \bibinfo{author}{Zhang,
  H.}, \bibinfo{author}{Wang, Z.}, \bibinfo{author}{Zuo, Z.},
  \bibinfo{author}{Li, A.}, \bibinfo{author}{Xing, W.}, \bibinfo{author}{Lu,
  D.}, \bibinfo{year}{2021}.
\newblock \bibinfo{title}{Diverse image style transfer via invertible
  cross-space mapping}, in: \bibinfo{booktitle}{2021 IEEE/CVF International
  Conference on Computer Vision (ICCV)}, \bibinfo{organization}{IEEE Computer
  Society}. pp. \bibinfo{pages}{14860--14869}.
%Type = Inproceedings
\bibitem[{Chiu and Gurari(2022)}]{photowct2}
\bibinfo{author}{Chiu, T.Y.}, \bibinfo{author}{Gurari, D.},
  \bibinfo{year}{2022}.
\newblock \bibinfo{title}{Photowct2: Compact autoencoder for photorealistic
  style transfer resulting from blockwise training and skip connections of
  high-frequency residuals}, in: \bibinfo{booktitle}{Proceedings of the
  IEEE/CVF Winter Conference on Applications of Computer Vision}, pp.
  \bibinfo{pages}{2868--2877}.
%Type = Article
\bibitem[{Dumoulin et~al.(2016)Dumoulin, Shlens and Kudlur}]{7}
\bibinfo{author}{Dumoulin, V.}, \bibinfo{author}{Shlens, J.},
  \bibinfo{author}{Kudlur, M.}, \bibinfo{year}{2016}.
\newblock \bibinfo{title}{A learned representation for artistic style}.
\newblock \bibinfo{journal}{arXiv preprint arXiv:1610.07629} .
%Type = Inproceedings
\bibitem[{Gatys et~al.(2016)Gatys, Ecker and Bethge}]{gatys}
\bibinfo{author}{Gatys, L.A.}, \bibinfo{author}{Ecker, A.S.},
  \bibinfo{author}{Bethge, M.}, \bibinfo{year}{2016}.
\newblock \bibinfo{title}{Image style transfer using convolutional neural
  networks}, in: \bibinfo{booktitle}{Proceedings of the IEEE conference on
  computer vision and pattern recognition}, pp. \bibinfo{pages}{2414--2423}.
%Type = Inproceedings
\bibitem[{Gu et~al.(2018)Gu, Chen, Liao and Yuan}]{10}
\bibinfo{author}{Gu, S.}, \bibinfo{author}{Chen, C.}, \bibinfo{author}{Liao,
  J.}, \bibinfo{author}{Yuan, L.}, \bibinfo{year}{2018}.
\newblock \bibinfo{title}{Arbitrary style transfer with deep feature
  reshuffle}, in: \bibinfo{booktitle}{Proceedings of the IEEE Conference on
  Computer Vision and Pattern Recognition}, pp. \bibinfo{pages}{8222--8231}.
%Type = Article
\bibitem[{He et~al.(2012)He, Sun and Tang}]{guide}
\bibinfo{author}{He, K.}, \bibinfo{author}{Sun, J.}, \bibinfo{author}{Tang,
  X.}, \bibinfo{year}{2012}.
\newblock \bibinfo{title}{Guided image filtering}.
\newblock \bibinfo{journal}{IEEE transactions on pattern analysis and machine
  intelligence} \bibinfo{volume}{35}, \bibinfo{pages}{1397--1409}.
%Type = Inproceedings
\bibitem[{Huang and Belongie(2017)}]{adain}
\bibinfo{author}{Huang, X.}, \bibinfo{author}{Belongie, S.},
  \bibinfo{year}{2017}.
\newblock \bibinfo{title}{Arbitrary style transfer in real-time with adaptive
  instance normalization}, in: \bibinfo{booktitle}{Proceedings of the IEEE
  international conference on computer vision}, pp.
  \bibinfo{pages}{1501--1510}.
%Type = Inproceedings
\bibitem[{Jing et~al.(2020)Jing, Liu, Ding, Wang, Ding, Song and
  Wen}]{DynamicIN}
\bibinfo{author}{Jing, Y.}, \bibinfo{author}{Liu, X.}, \bibinfo{author}{Ding,
  Y.}, \bibinfo{author}{Wang, X.}, \bibinfo{author}{Ding, E.},
  \bibinfo{author}{Song, M.}, \bibinfo{author}{Wen, S.}, \bibinfo{year}{2020}.
\newblock \bibinfo{title}{Dynamic instance normalization for arbitrary style
  transfer}, in: \bibinfo{booktitle}{Proceedings of the AAAI Conference on
  Artificial Intelligence}, pp. \bibinfo{pages}{4369--4376}.
%Type = Inproceedings
\bibitem[{Jing et~al.(2018)Jing, Liu, Yang, Feng, Yu, Tao and Song}]{stroke}
\bibinfo{author}{Jing, Y.}, \bibinfo{author}{Liu, Y.}, \bibinfo{author}{Yang,
  Y.}, \bibinfo{author}{Feng, Z.}, \bibinfo{author}{Yu, Y.},
  \bibinfo{author}{Tao, D.}, \bibinfo{author}{Song, M.}, \bibinfo{year}{2018}.
\newblock \bibinfo{title}{Stroke controllable fast style transfer with adaptive
  receptive fields}, in: \bibinfo{booktitle}{Proceedings of the European
  Conference on Computer Vision (ECCV)}, pp. \bibinfo{pages}{238--254}.
%Type = Inproceedings
\bibitem[{Johnson et~al.(2016)Johnson, Alahi and Fei-Fei}]{Perceptual}
\bibinfo{author}{Johnson, J.}, \bibinfo{author}{Alahi, A.},
  \bibinfo{author}{Fei-Fei, L.}, \bibinfo{year}{2016}.
\newblock \bibinfo{title}{Perceptual losses for real-time style transfer and
  super-resolution}, in: \bibinfo{booktitle}{Computer Vision--ECCV 2016: 14th
  European Conference, Amsterdam, The Netherlands, October 11-14, 2016,
  Proceedings, Part II 14}, \bibinfo{organization}{Springer}. pp.
  \bibinfo{pages}{694--711}.
%Type = Article
\bibitem[{Kingma and Ba(2014)}]{adam}
\bibinfo{author}{Kingma, D.P.}, \bibinfo{author}{Ba, J.}, \bibinfo{year}{2014}.
\newblock \bibinfo{title}{Adam: A method for stochastic optimization}.
\newblock \bibinfo{journal}{arXiv preprint arXiv:1412.6980} .
%Type = Inproceedings
\bibitem[{Kolkin et~al.(2019)Kolkin, Salavon and Shakhnarovich}]{15}
\bibinfo{author}{Kolkin, N.}, \bibinfo{author}{Salavon, J.},
  \bibinfo{author}{Shakhnarovich, G.}, \bibinfo{year}{2019}.
\newblock \bibinfo{title}{Style transfer by relaxed optimal transport and
  self-similarity}, in: \bibinfo{booktitle}{Proceedings of the IEEE/CVF
  Conference on Computer Vision and Pattern Recognition}, pp.
  \bibinfo{pages}{10051--10060}.
%Type = Inproceedings
\bibitem[{Li and Wand(2016a)}]{17}
\bibinfo{author}{Li, C.}, \bibinfo{author}{Wand, M.}, \bibinfo{year}{2016}a.
\newblock \bibinfo{title}{Combining markov random fields and convolutional
  neural networks for image synthesis}, in: \bibinfo{booktitle}{Proceedings of
  the IEEE conference on computer vision and pattern recognition}, pp.
  \bibinfo{pages}{2479--2486}.
%Type = Inproceedings
\bibitem[{Li and Wand(2016b)}]{18}
\bibinfo{author}{Li, C.}, \bibinfo{author}{Wand, M.}, \bibinfo{year}{2016}b.
\newblock \bibinfo{title}{Precomputed real-time texture synthesis with
  markovian generative adversarial networks}, in: \bibinfo{booktitle}{Computer
  Vision--ECCV 2016: 14th European Conference, Amsterdam, The Netherlands,
  October 11-14, 2016, Proceedings, Part III 14},
  \bibinfo{organization}{Springer}. pp. \bibinfo{pages}{702--716}.
%Type = Inproceedings
\bibitem[{Li et~al.(2017a)Li, Fang, Yang, Wang, Lu and Yang}]{20}
\bibinfo{author}{Li, Y.}, \bibinfo{author}{Fang, C.}, \bibinfo{author}{Yang,
  J.}, \bibinfo{author}{Wang, Z.}, \bibinfo{author}{Lu, X.},
  \bibinfo{author}{Yang, M.H.}, \bibinfo{year}{2017}a.
\newblock \bibinfo{title}{Diversified texture synthesis with feed-forward
  networks}, in: \bibinfo{booktitle}{Proceedings of the IEEE conference on
  computer vision and pattern recognition}, pp. \bibinfo{pages}{3920--3928}.
%Type = Article
\bibitem[{Li et~al.(2017b)Li, Fang, Yang, Wang, Lu and Yang}]{21}
\bibinfo{author}{Li, Y.}, \bibinfo{author}{Fang, C.}, \bibinfo{author}{Yang,
  J.}, \bibinfo{author}{Wang, Z.}, \bibinfo{author}{Lu, X.},
  \bibinfo{author}{Yang, M.H.}, \bibinfo{year}{2017}b.
\newblock \bibinfo{title}{Universal style transfer via feature transforms}.
\newblock \bibinfo{journal}{Advances in neural information processing systems}
  \bibinfo{volume}{30}.
%Type = Inproceedings
\bibitem[{Li et~al.(2018)Li, Liu, Li, Yang and Kautz}]{closed}
\bibinfo{author}{Li, Y.}, \bibinfo{author}{Liu, M.Y.}, \bibinfo{author}{Li,
  X.}, \bibinfo{author}{Yang, M.H.}, \bibinfo{author}{Kautz, J.},
  \bibinfo{year}{2018}.
\newblock \bibinfo{title}{A closed-form solution to photorealistic image
  stylization}, in: \bibinfo{booktitle}{Proceedings of the European conference
  on computer vision (ECCV)}, pp. \bibinfo{pages}{453--468}.
%Type = Article
\bibitem[{Li et~al.(2017c)Li, Wang, Liu and Hou}]{demystifying}
\bibinfo{author}{Li, Y.}, \bibinfo{author}{Wang, N.}, \bibinfo{author}{Liu,
  J.}, \bibinfo{author}{Hou, X.}, \bibinfo{year}{2017}c.
\newblock \bibinfo{title}{Demystifying neural style transfer}.
\newblock \bibinfo{journal}{arXiv preprint arXiv:1701.01036} .
%Type = Inproceedings
\bibitem[{Lin et~al.(2014)Lin, Maire, Belongie, Hays, Perona, Ramanan,
  Doll{\'a}r and Zitnick}]{coco}
\bibinfo{author}{Lin, T.Y.}, \bibinfo{author}{Maire, M.},
  \bibinfo{author}{Belongie, S.}, \bibinfo{author}{Hays, J.},
  \bibinfo{author}{Perona, P.}, \bibinfo{author}{Ramanan, D.},
  \bibinfo{author}{Doll{\'a}r, P.}, \bibinfo{author}{Zitnick, C.L.},
  \bibinfo{year}{2014}.
\newblock \bibinfo{title}{Microsoft coco: Common objects in context}, in:
  \bibinfo{booktitle}{Computer Vision--ECCV 2014: 13th European Conference,
  Zurich, Switzerland, September 6-12, 2014, Proceedings, Part V 13},
  \bibinfo{organization}{Springer}. pp. \bibinfo{pages}{740--755}.
%Type = Inproceedings
\bibitem[{Luan et~al.(2017)Luan, Paris, Shechtman and Bala}]{deep}
\bibinfo{author}{Luan, F.}, \bibinfo{author}{Paris, S.},
  \bibinfo{author}{Shechtman, E.}, \bibinfo{author}{Bala, K.},
  \bibinfo{year}{2017}.
\newblock \bibinfo{title}{Deep photo style transfer}, in:
  \bibinfo{booktitle}{Proceedings of the IEEE conference on computer vision and
  pattern recognition}, pp. \bibinfo{pages}{4990--4998}.
%Type = Inproceedings
\bibitem[{Park and Lee(2019)}]{SANet}
\bibinfo{author}{Park, D.Y.}, \bibinfo{author}{Lee, K.H.},
  \bibinfo{year}{2019}.
\newblock \bibinfo{title}{Arbitrary style transfer with style-attentional
  networks}, in: \bibinfo{booktitle}{proceedings of the IEEE/CVF conference on
  computer vision and pattern recognition}, pp. \bibinfo{pages}{5880--5888}.
%Type = Article
\bibitem[{Phillips and Mackintosh(2011)}]{wiki}
\bibinfo{author}{Phillips, F.}, \bibinfo{author}{Mackintosh, B.},
  \bibinfo{year}{2011}.
\newblock \bibinfo{title}{Wiki art gallery, inc.: A case for critical
  thinking}.
\newblock \bibinfo{journal}{Issues in Accounting Education}
  \bibinfo{volume}{26}, \bibinfo{pages}{593--608}.
%Type = Inproceedings
\bibitem[{Pitie et~al.(2005)Pitie, Kokaram and Dahyot}]{pitie2005n}
\bibinfo{author}{Pitie, F.}, \bibinfo{author}{Kokaram, A.C.},
  \bibinfo{author}{Dahyot, R.}, \bibinfo{year}{2005}.
\newblock \bibinfo{title}{N-dimensional probability density function transfer
  and its application to color transfer}, in: \bibinfo{booktitle}{Tenth IEEE
  International Conference on Computer Vision (ICCV'05) Volume 1},
  \bibinfo{organization}{IEEE}. pp. \bibinfo{pages}{1434--1439}.
%Type = Article
\bibitem[{Piti{\'e} et~al.(2007)Piti{\'e}, Kokaram and
  Dahyot}]{pitie2007automated}
\bibinfo{author}{Piti{\'e}, F.}, \bibinfo{author}{Kokaram, A.C.},
  \bibinfo{author}{Dahyot, R.}, \bibinfo{year}{2007}.
\newblock \bibinfo{title}{Automated colour grading using colour distribution
  transfer}.
\newblock \bibinfo{journal}{Computer Vision and Image Understanding}
  \bibinfo{volume}{107}, \bibinfo{pages}{123--137}.
%Type = Article
\bibitem[{Reinhard et~al.(2001)Reinhard, Adhikhmin, Gooch and
  Shirley}]{reinhard2001color}
\bibinfo{author}{Reinhard, E.}, \bibinfo{author}{Adhikhmin, M.},
  \bibinfo{author}{Gooch, B.}, \bibinfo{author}{Shirley, P.},
  \bibinfo{year}{2001}.
\newblock \bibinfo{title}{Color transfer between images}.
\newblock \bibinfo{journal}{IEEE Computer graphics and applications}
  \bibinfo{volume}{21}, \bibinfo{pages}{34--41}.
%Type = Article
\bibitem[{Risser et~al.(2017)Risser, Wilmot and Barnes}]{27}
\bibinfo{author}{Risser, E.}, \bibinfo{author}{Wilmot, P.},
  \bibinfo{author}{Barnes, C.}, \bibinfo{year}{2017}.
\newblock \bibinfo{title}{Stable and controllable neural texture synthesis and
  style transfer using histogram losses}.
\newblock \bibinfo{journal}{arXiv preprint arXiv:1701.08893} .
%Type = Article
\bibitem[{Sengupta et~al.(2019)Sengupta, Ye, Wang, Liu and Roy}]{vgg}
\bibinfo{author}{Sengupta, A.}, \bibinfo{author}{Ye, Y.},
  \bibinfo{author}{Wang, R.}, \bibinfo{author}{Liu, C.}, \bibinfo{author}{Roy,
  K.}, \bibinfo{year}{2019}.
\newblock \bibinfo{title}{Going deeper in spiking neural networks: Vgg and
  residual architectures}.
\newblock \bibinfo{journal}{Frontiers in neuroscience} \bibinfo{volume}{13},
  \bibinfo{pages}{95}.
%Type = Inproceedings
\bibitem[{Shen et~al.(2018)Shen, Yan and Zeng}]{meta}
\bibinfo{author}{Shen, F.}, \bibinfo{author}{Yan, S.}, \bibinfo{author}{Zeng,
  G.}, \bibinfo{year}{2018}.
\newblock \bibinfo{title}{Neural style transfer via meta networks}, in:
  \bibinfo{booktitle}{Proceedings of the IEEE Conference on Computer Vision and
  Pattern Recognition}, pp. \bibinfo{pages}{8061--8069}.
%Type = Article
\bibitem[{ShiQi~Jiang(2023a)}]{CTDP}
\bibinfo{author}{ShiQi~Jiang, JunJie~Kang, Y.L.}, \bibinfo{year}{2023}a.
\newblock \bibinfo{title}{Color and texture dual pipeline lightweight style
  transfer}.
\newblock \bibinfo{journal}{arXiv preprint arXiv:2310.01321} .
%Type = Article
\bibitem[{ShiQi~Jiang(2023b)}]{DcDae}
\bibinfo{author}{ShiQi~Jiang, JunJie~Kang, Y.L.}, \bibinfo{year}{2023}b.
\newblock \bibinfo{title}{Degree-controllable lightweight fast style transfer
  with detail attention-enhanced}.
\newblock \bibinfo{journal}{arXiv preprint arXiv:2306.16846} .
%Type = Article
\bibitem[{Ulyanov et~al.(2016a)Ulyanov, Lebedev, Vedaldi and Lempitsky}]{32}
\bibinfo{author}{Ulyanov, D.}, \bibinfo{author}{Lebedev, V.},
  \bibinfo{author}{Vedaldi, A.}, \bibinfo{author}{Lempitsky, V.},
  \bibinfo{year}{2016}a.
\newblock \bibinfo{title}{Texture networks: Feed-forward synthesis of textures
  and stylized images}.
\newblock \bibinfo{journal}{arXiv preprint arXiv:1603.03417} .
%Type = Article
\bibitem[{Ulyanov et~al.(2016b)Ulyanov, Vedaldi and Lempitsky}]{33}
\bibinfo{author}{Ulyanov, D.}, \bibinfo{author}{Vedaldi, A.},
  \bibinfo{author}{Lempitsky, V.}, \bibinfo{year}{2016}b.
\newblock \bibinfo{title}{Instance normalization: The missing ingredient for
  fast stylization}.
\newblock \bibinfo{journal}{arXiv preprint arXiv:1607.08022} .
%Type = Inproceedings
\bibitem[{Wang et~al.(2020)Wang, Li, Wang, Hu and Yang}]{collaborative}
\bibinfo{author}{Wang, H.}, \bibinfo{author}{Li, Y.}, \bibinfo{author}{Wang,
  Y.}, \bibinfo{author}{Hu, H.}, \bibinfo{author}{Yang, M.H.},
  \bibinfo{year}{2020}.
\newblock \bibinfo{title}{Collaborative distillation for ultra-resolution
  universal style transfer}, in: \bibinfo{booktitle}{Proceedings of the
  IEEE/CVF conference on computer vision and pattern recognition}, pp.
  \bibinfo{pages}{1860--1869}.
%Type = Inproceedings
\bibitem[{Wang et~al.(2017)Wang, Oxholm, Zhang and Wang}]{35}
\bibinfo{author}{Wang, X.}, \bibinfo{author}{Oxholm, G.},
  \bibinfo{author}{Zhang, D.}, \bibinfo{author}{Wang, Y.F.},
  \bibinfo{year}{2017}.
\newblock \bibinfo{title}{Multimodal transfer: A hierarchical deep
  convolutional neural network for fast artistic style transfer}, in:
  \bibinfo{booktitle}{Proceedings of the IEEE conference on computer vision and
  pattern recognition}, pp. \bibinfo{pages}{5239--5247}.
%Type = Article
\bibitem[{Wang et~al.(2021)Wang, Zhao, Chen, Zuo, Li, Xing and Lu}]{wang}
\bibinfo{author}{Wang, Z.}, \bibinfo{author}{Zhao, L.}, \bibinfo{author}{Chen,
  H.}, \bibinfo{author}{Zuo, Z.}, \bibinfo{author}{Li, A.},
  \bibinfo{author}{Xing, W.}, \bibinfo{author}{Lu, D.}, \bibinfo{year}{2021}.
\newblock \bibinfo{title}{Divswapper: towards diversified patch-based arbitrary
  style transfer}.
\newblock \bibinfo{journal}{arXiv preprint arXiv:2101.06381} .
%Type = Article
\bibitem[{Wang et~al.(2022)Wang, Zhao, Zuo, Li, Chen, Xing and Lu}]{microast}
\bibinfo{author}{Wang, Z.}, \bibinfo{author}{Zhao, L.}, \bibinfo{author}{Zuo,
  Z.}, \bibinfo{author}{Li, A.}, \bibinfo{author}{Chen, H.},
  \bibinfo{author}{Xing, W.}, \bibinfo{author}{Lu, D.}, \bibinfo{year}{2022}.
\newblock \bibinfo{title}{Microast: Towards super-fast ultra-resolution
  arbitrary style transfer}.
\newblock \bibinfo{journal}{arXiv preprint arXiv:2211.15313} .
%Type = Inproceedings
\bibitem[{Wen et~al.(2023)Wen, Gao and Zou}]{cap}
\bibinfo{author}{Wen, L.}, \bibinfo{author}{Gao, C.}, \bibinfo{author}{Zou,
  C.}, \bibinfo{year}{2023}.
\newblock \bibinfo{title}{Cap-vstnet: Content affinity preserved versatile
  style transfer}, in: \bibinfo{booktitle}{Proceedings of the IEEE/CVF
  Conference on Computer Vision and Pattern Recognition}, pp.
  \bibinfo{pages}{18300--18309}.
%Type = Inproceedings
\bibitem[{Xie et~al.(2022)Xie, Li, Huang, Fu, Wang and Guo}]{xie}
\bibinfo{author}{Xie, X.}, \bibinfo{author}{Li, Y.}, \bibinfo{author}{Huang,
  H.}, \bibinfo{author}{Fu, H.}, \bibinfo{author}{Wang, W.},
  \bibinfo{author}{Guo, Y.}, \bibinfo{year}{2022}.
\newblock \bibinfo{title}{Artistic style discovery with independent
  components}, in: \bibinfo{booktitle}{Proceedings of the IEEE/CVF Conference
  on Computer Vision and Pattern Recognition}, pp.
  \bibinfo{pages}{19870--19879}.
%Type = Inproceedings
\bibitem[{Yoo et~al.(2019)Yoo, Uh, Chun, Kang and Ha}]{photorealistic}
\bibinfo{author}{Yoo, J.}, \bibinfo{author}{Uh, Y.}, \bibinfo{author}{Chun,
  S.}, \bibinfo{author}{Kang, B.}, \bibinfo{author}{Ha, J.W.},
  \bibinfo{year}{2019}.
\newblock \bibinfo{title}{Photorealistic style transfer via wavelet
  transforms}, in: \bibinfo{booktitle}{Proceedings of the IEEE/CVF
  International Conference on Computer Vision}, pp.
  \bibinfo{pages}{9036--9045}.
%Type = Inproceedings
\bibitem[{Zhang et~al.(2019)Zhang, Zhu and Zhu}]{zhang}
\bibinfo{author}{Zhang, C.}, \bibinfo{author}{Zhu, Y.}, \bibinfo{author}{Zhu,
  S.C.}, \bibinfo{year}{2019}.
\newblock \bibinfo{title}{Metastyle: Three-way trade-off among speed,
  flexibility, and quality in neural style transfer}, in:
  \bibinfo{booktitle}{Proceedings of the AAAI Conference on Artificial
  Intelligence}, pp. \bibinfo{pages}{1254--1261}.
%Type = Inproceedings
\bibitem[{Zhang and Dana(2018a)}]{37}
\bibinfo{author}{Zhang, H.}, \bibinfo{author}{Dana, K.}, \bibinfo{year}{2018}a.
\newblock \bibinfo{title}{Multi-style generative network for real-time
  transfer}, in: \bibinfo{booktitle}{Proceedings of the European Conference on
  Computer Vision (ECCV) Workshops}, pp. \bibinfo{pages}{0--0}.
%Type = Inproceedings
\bibitem[{Zhang and Dana(2018b)}]{12}
\bibinfo{author}{Zhang, H.}, \bibinfo{author}{Dana, K.}, \bibinfo{year}{2018}b.
\newblock \bibinfo{title}{Multi-style generative network for real-time
  transfer}, in: \bibinfo{booktitle}{Proceedings of the European Conference on
  Computer Vision (ECCV) Workshops}, pp. \bibinfo{pages}{0--0}.

\end{thebibliography}

\end{document}